# Mapping Instructors’ Attitudes Towards Cheating with AI: An Exploratory Analysis Using Forum Analysis

Iman Mohammadi, Ali Keramati, Sina Rismanchian
imohamm1@uci.edu, a.kera@uci.edu, srismanc@uci.edu
University of California, Irvine

**Abstract:** In teaching and learning environments, cheating and the violation of academic integrity have always been one of the challenges, involving different levels of concerns such as administrative, personal knowledge, learning, and equity. The introduction of large language models has brought cheating in dialogue to the forefront, given the apparent capabilities of these models to complete students’ academic assignments. This study aims to leverage recent dialogues among university professors about cheating with Generative AI to better understand their beliefs about academic honesty. In doing so, we analyze a corpus of 2.01 million posts (threads and comments) in the r/Professors subreddit discussing cheating after the release of ChatGPT, providing a better understanding of teachers’ motivations to discourage cheating and the strategies they use to do so. We synthesize the why and how of addressing cheating to identify caveats about the future of AI and academic integration.

## Objectives and Significance

After the introduction of large language models (LLMs) and generative artificial intelligence (GenAI) chatbots, the idea of students using these tools extensively to bypass the effort required to complete academic assignments, under the guise of cheating, academic dishonesty, etc., has made headlines (Yusuf, 2024). Cheating and preaching academic integrity are not new phenomena in higher education and K-12 settings (Reddy-Lund, 2025); however, with the broad adoption of these tools, cheating has resurfaced (von Garrel, 2023). Several research projects have been started to identify the level of cheating among students (Sozon, 2024; Rismanchian et al., 2026). On the other hand, many professors have reported taking the initiative to fight cheating with GenAI (Kazley, 2024). Although there is an ongoing effort to better understand how widespread cheating with AI is among students, professors' reaction to the perceived broad usage can change the way they design their instructional policies, their course assignments, and more broadly, the dynamic of teaching and learning might enter into a new era (Sun, 2025; Sumakul, 2022). From the lens of the learning sciences, it is important to understand how these transitions to an era in which GenAI is accessible to all students might or might not align with the learning sciences' epistemology, theories, and praxis (Balabdaoui et al., 2024).

This research aims to provide a clearer picture of teachers’ motivations and strategies for addressing students’ cheating and irresponsible use of AI, and to offer a few considerations from a learning sciences perspective that might help better understand how the dynamics of teaching and learning in higher education are evolving.

University faculty are currently engaging in sense-making in the open about these shifts, especially in public-practitioner venues. These venues offer a naturally occurring record of how instructors narrate the why (motivations and values) and the how (design and enforcement moves) of academic honesty in the era of GenAI. Mapping these narratives matters for the learning sciences because they reveal the design logics educators are settling on, whether toward surveillance and control, toward redesign for learning, or some hybrid, and the equity/ethics trade-offs implied by each.

Building on recent concerns about AI-enabled cheating and the ongoing sense-making among instructors, this study addresses the following research questions:

- **RQ1:** How do instructors describe their motivations and values in responding to AI-mediated cheating?
- **RQ2:** What strategies do instructors report using to address or prevent AI-mediated cheating in their courses?

This paper contributes: 1) Thematic account of instructors’ motivations for combating AI-mediated cheating (the "why"); 2) A synthesis of methods they report using (the "how"), spanning pedagogical redesign, technological controls, and communication strategies; and 3) Implications that align (or clash) with learning-sciences commitments to authenticity, equity, participation.

This study is informed by learning sciences perspectives that conceptualize learning as participation in meaningful disciplinary activity rather than the production of decontextualized outputs (Lave & Wenger, 1991; Wenger, 1998). From this perspective, concerns about AI-mediated cheating can be understood not only as violations of academic rules but as disruptions to how learning becomes visible through student engagement in disciplinary practices (Shepard, 2000). This framing guides how we interpret instructors' motivations and strategies in responding to AI use in their classrooms.

## Methodology

To address this gap in instructors' perceptions, we used Reddit, one of the most popular online communities, as our data source. Reddit provides a public space where university professors and instructors share their thoughts, experiences, and challenges in teaching. Specifically, we analyzed posts and comments from the r/Professors subreddit, which is a public forum where educators discuss academic life and pedagogy. This platform was chosen because it allows open, candid reflections from instructors about emerging issues such as AI use and cheating.

### Data Collection

To collect data, we used Reddit's public API and keyword-based search to identify posts and discussions relevant to academic integrity and AI use. The keyword list included terms related to academic honesty, integrity, and misconduct, as well as AI tools and policies (e.g., ChatGPT, AI detection, Turnitin, plagiarism, Grammarly, GPTZero, GenAI, LLM, proctoring, AI policy, academic conduct).

We searched for posts made after the release of ChatGPT in late 2022, covering a period when discussions about AI cheating became widespread. This initial search yielded around 2.01 million posts (threads and comments combined) from r/Professors.

### Sampling

Because the dataset was large, we applied random sampling to select a manageable subset of posts for analysis. Sampling was done to ensure representation across different time periods and discussion contexts, so that posts from early reactions to more recent reflections were both included. In this process, we collected a sample of 2000 combined threads and comments, then analyzed them.

### Data Analysis

We used qualitative thematic analysis (Braun & Clarke, 2006) to identify recurring ideas, rationales, and strategies expressed by instructors. First, we performed open coding on a sample of posts to capture initial ideas. Then, through iterative coding and memoing, we refined these into higher-level themes and subthemes. The focus was twofold:

1. Why do instructors fight or respond to AI-enabled cheating?
2. How do they act to address it in their teaching practice?

The analysis was done manually by the authors to ensure interpretive depth and sensitivity to context.

## Results

Upon completing our thematic analysis of the personal reasons instructors use to fight AI-enabled cheating and their strategies, we identified three general themes for each question.

### Why do instructors fight cheating?

From the exploratory analysis of posts and comments, three themes of personal reasons, beyond institutional considerations such as complying with university policies, emerged as reasons professors state to be their reasons for fighting cheating: 1) Protecting students' learning process, 2) Professors' well-being, and 3) Maintaining the integrity of their values in their instructional practice. Regarding the first theme, a professor mentioned that:

> I'm hearing from students, usually in disciplines outside of mine in the gen eds I teach, that they use ChatGPT to get over the "difficult" parts of the writing **process,** whether it be coming up with a topic to write about, connecting ideas together, summarizing a source, or writing an intro and conclusion to their essay. Even if they are just using ChatGPT to "get going," students miss out on exercising their creative-invention skills. If a student is only doing the easy parts of the writing **process**, or relying on something (or someone else) to do it for them, then they are not learning."

Many similar anecdotes from professors underscore the importance of the learning process and the ways they believe students' learning can be undermined when they overrely on these tools. Furthermore, professors mention frustration over their overwhelming interactions with AI-generated essays, which have removed the personal connection between them and their students, leading to lower satisfaction with their jobs and mental well-being:

> There's no dialogue, and it makes me really f[...] sad. AI did not just change how writing works. It has completely changed the classroom atmosphere. Students are suspicious of me and see me as nothing but a possible obstacle, and they won't even answer if I ask how they are doing.

Or in another similar situation, a professor mentions that: "I grew sick and tired of wondering if what I graded was written by a human or not", so they decided to shift their assignments to "handwritten" essays. Finally, professors mention values, such as integrity and equity, as reasons to fight cheating, which might in turn create ethical dilemmas for the instructors themselves, as a professor posted:

> I'm struggling with the tension between promoting equity and maintaining academic integrity. I wholeheartedly support inclusive, student-centered education. But in practice, I see widespread academic dishonesty such as AI misuse, ghostwriting, and test cheating, with little appetite from institutions to address it meaningfully.

## How do instructors address cheating?

Among those professors who act upon cheating, there are three broad themes for their techniques: 1) Pedagogical and assessment redesign, 2) Technological control, and 3) Communication. We will provide details on each of these categories:

### Pedagogical and assessment redesign

Educators redesign tasks and assessment ecologies to make unauthorized AI use less useful and learning more visible. For instance, De-AI-ing through analog and in-person methods: Handwritten, supervised, and paper-based work (often with reduced load) to ensure authorship and focus: "Everything I grade… will be handwritten… write in-class essays using the notes." Strategies include withholding quiz copies to limit item leakage and attending to small operational details (e.g., bathroom breaks) when integrity risk is high.

Process-oriented and scaffolded assessment is also a prevalent strategy, which means shifting weight from final products to steps (in-class invention, annotated sources, screenshot trails, iterative drafts), both deters outsourcing and evidence learning:

> I'm thinking of shifting all my points in class to be on the processes of writing (topic invention done in class, drafts maybe done by hand or with recorded proof of the process like taking a few screenshots along the way, collecting sources, annotations, etc.) and making my final papers equivalent to next-to-nothing

Oral checks and process interviews are used to investigate authorship, alongside task tweaks demanding primary sources and critical application. And finally, Creative "bait" questions: Designing items that require local course knowledge or personal context, with selective "off-syllabus" choices to catch inattentive AI use: philosophy short-answer sets where only some options were covered in class.

### Technological control

Monitoring and deterrence tactics are deployed with varying comfort and skepticism, such as proctoring software and in-person monitoring, combinations of lockdown browsers, timed windows, and identity checks. Some instructors also report using detectors as triage signals: "Two different ChatGPT detectors show it was written by AI… the cheating jumps off the page even without the detector."

Furthermore, professors reported using tools such as GPTZero, which also appear to control students' writing when they write their essays.

### Communication

The last theme speaks to the instructor's attempt to promote dialogues about the responsible use of AI, how it might impact their learning, and even scare students about how an instructor can use prompt engineering techniques, such as hidden prompt injection, to identify students who used AI. For instance, an instructor mentioned:

> We also discussed the ethical implications of using GPT in various scenarios. Just for fun, I had GPT write a love letter to a fictional partner and then asked whether they would be okay

> with receiving it from their GF or BF. Unsurprisingly, most did not think it was ok. (getting them to think a little about ethical implications…)

Or, in another case, a professor opts to scare students about the possibility of tracing students' AI usage:

> then opened a Word document on my screen, which showed absolutely nothing... I dragged this file into ChatGPT and asked it to summarize the document. ChatGPT began generating a full-page summary of my course syllabus. (This was achieved by using the 'hidden' text feature in Word). I explained to the students that there are many ways AI tools could leave a trace…

**Table 1**
*Thematic Framework of Instructors' Motivations and Strategies for Combating AI-Mediated Cheating*

| RQs | Themes | Description |
|---|---|---|
| | 1A. Protecting Student Learning. | Instructors express concern that reliance on AI undermines students' ability to think, create, and learn through effort. |
| 1. Motivations for Combating AI Cheating | 1B. Instructor Well-being and Professional Fulfillment | Repeated encounters with AI-generated work lead to frustration and diminish teachers' sense of connection and purpose. |
| | 1C. Upholding Values of Integrity and Equity | Teachers feel a moral duty to maintain fairness and honesty, even when doing so creates ethical dilemmas or emotional strain. |
| | | |
| | 2A. Pedagogical and Assessment Redesign | Educators redesign tasks and assessment ecologies to make unauthorized AI use less useful and learning more visible |
| 2. Strategies for Addressing AI Cheating | 2B. Technological Control | Using proctoring tools, lockdown browsers, or AI detectors as deterrents or verification measures. |
| | 2C. Communication and Ethical Dialogue | Promoting discussions with students about responsible AI use, ethics, and the importance of authentic learning. |

Our analysis highlights two broad categories of instructor responses to AI-mediated cheating: their motivations for acting and the strategies they employ.

## Discussion and Conclusion

This study demonstrates the value of analyzing open-ended forum data to capture how instructors actively engage in sense-making about AI-mediated cheating and adapt their practices in real time. Our findings provide insight into both the challenges instructors face and the diverse strategies they employ, without positioning these responses as inherently right or wrong. Instead, we interpret these patterns through a learning sciences lens that conceptualizes learning as participation in meaningful disciplinary activity rather than the production of decontextualized outputs (Lave & Wenger, 1991; Wenger, 1998). From this perspective, instructors' concerns about AI-mediated cheating are not only about rule violations, but about students' disengagement from the processes through which learning becomes visible, including the effortful construction and articulation of knowledge (Scardamalia & Bereiter, 2014).

At the same time, our findings reveal important variation and tension in how instructors respond. Some strategies, such as process-oriented assessment and constructive communication, can be understood as attempts to re-center learning around participation, reflection, and student agency, aligning with assessment perspectives that emphasize learning-oriented environments (Shepard, 2000). These approaches help shift conversations about academic integrity from compliance toward shared understanding and responsibility. In contrast, other strategies, particularly those focused on surveillance, detection, and strict control, prioritize verification of authorship and compliance, and may be less aligned with these principles (Black & Wiliam, 1998). While such approaches may address immediate concerns about misuse, they risk narrowing what counts as evidence of learning to what can be monitored or verified. Rather than evaluating these approaches normatively, we highlight this tension as a productive space for inquiry. AI-mediated cheating thus emerges not only as a challenge to academic integrity but as a catalyst for reconsidering what counts as evidence of learning, how authenticity is defined, and how fairness is enacted in contemporary educational contexts.

The results of our analysis show that understanding instructors' motivations and strategies for responding to AI-mediated cheating is essential for aligning future academic integrity practices with learning sciences commitments to authenticity, equity, and meaningful participation.